\documentclass{article} 
\usepackage{iclr2026_conference,times}

\usepackage{amsmath,amsfonts,bm}

\def\eqref#1{equation~\ref{#1}}

\def\1{\bm{1}}

\DeclareMathAlphabet{\mathsfit}{\encodingdefault}{\sfdefault}{m}{sl}
\SetMathAlphabet{\mathsfit}{bold}{\encodingdefault}{\sfdefault}{bx}{n}

\usepackage{hyperref}
\usepackage{url}
\usepackage{booktabs} 
\usepackage{graphicx}
\usepackage{multirow}
\graphicspath{{./img/}}

\title{Spectral Gaps and Spatial Priors: Studying Hyperspectral Downstream Adaptation Using TerraMind}

\author{Julia A. Leonardi \& Maria A. Brovelli 
\\
Department of Civil and Environmental Engineering\\
Politecnico di Milano\\
Milan, Italy \\
\texttt{\{juliaanna.leonardi, maria.brovelli\}@polimi.it} \\
\And
Johannes Jakubik \& Paolo Fraccaro \\
IBM Research Europe \\
Rüschlikon, Switzerland \\
\texttt{\{johannes.jakubik1, paolo.fraccaro\}@ibm.com} \\
}

\iclrfinalcopy 
\begin{document}

\maketitle

\begin{abstract}
Geospatial Foundation Models (GFMs) typically lack native support for Hyperspectral Imaging (HSI) due to the complexity and sheer size of high-dimensional spectral data. This study investigates the adaptability of TerraMind, a multimodal GFM, to HSI downstream tasks \emph{without} HSI-specific pretraining by comparing two channel adaptation strategies: Naive Band Selection and physics-aware Spectral Response Function (SRF) grouping. Our results confirm the general superiority of HSI-native architectures, though TerraMind demonstrates robust adaptability to spectral tasks through simple band selection. By establishing this baseline, we underscore the necessity of developing native spectral tokenization for future multimodal GFMs. \looseness=-1
\end{abstract}

\section{Introduction}
Geospatial Foundation Models (GFMs) have emerged as a paradigm shift in remote sensing, moving from task-specific architectures to generalist models pretrained on vast datasets. GFMs aim to learn scalable, transferable representations across diverse downstream tasks, which are generalizable to a global scale. However, remote sensing presents unique challenges due to high heterogeneity in sensor modalities. Although modern GFMs can incorporate diverse input types, from optical imagery to SAR, Hyperspectral Imaging (HSI), which is essential for precision agriculture, mineral exploration, and environmental monitoring, remains underrepresented. HSI offers significant spectral detail through hundreds of narrow spectral channels, but poses challenges in data complexity. Although the availability of HSI data is increasing via satellite missions like EnMAP \citep{Kaufmann_2012} and PRISMA \citep{Guarini_2018}, Deep Learning (DL) applications remain task-specific \citep{Ibanez_2022, Scheibenreif_2023, Xu_2025}. Recent works have introduced HSI-specific GFMs, such as HyperSIGMA \citep{Wang_2024} or SpectralEarth \citep{Braham_2024}. However, these are largely unimodal. Current multimodal GFMs that incorporate HSI into the pretraining data (e.g., DOFA \citep{Xiong_2024}) rarely address the distinct three-dimensional feature extraction required for HSI, thereby creating a gap in effective multimodal integration.  An additional challenge concerns the limited availability and diversity of downstream datasets suited for DL applications, with most benchmark HSI datasets being single scenes, not allowing for a proper training process and hindering generalization \citep{Signoroni_2019}. In this work, we address whether existing multimodal GFMs, pretrained without HSI, can serve as effective baselines for HSI-specific tasks. \looseness=-1

\begin{figure}[h]
    \centering
    \includegraphics[width=\linewidth]{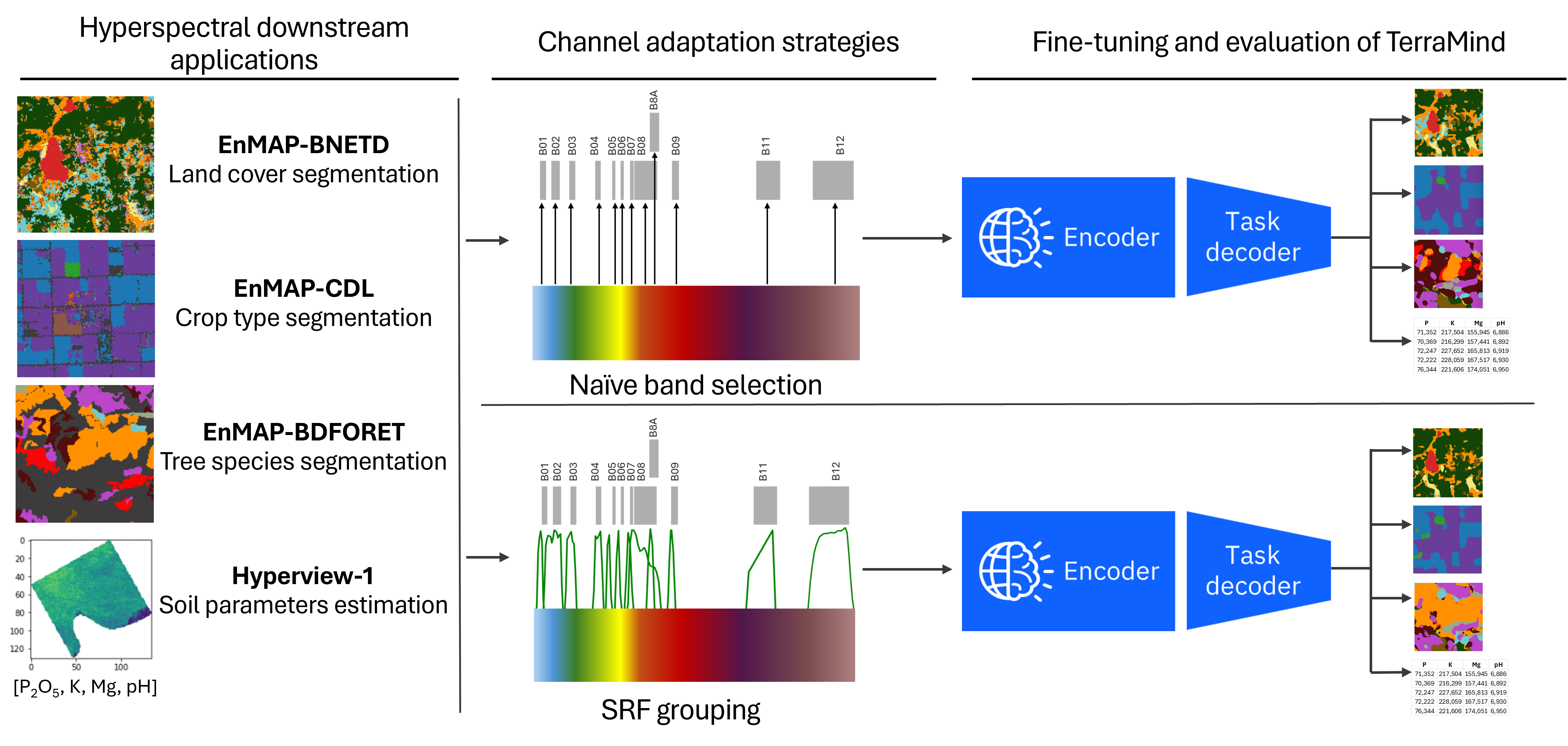}
    \caption{Experimental framework for benchmarking channel adaptation strategies (Naive Selection vs. SRF Grouping) using the TerraMind backbone on four HSI-specific applications.}
    \label{fig:vis}
\end{figure}

We fine-tune TerraMind \citep{Jakubik_2025}, a multimodal GFM, on four different HSI downstream tasks. To bridge the modality gap, we compare two channel adaptation strategies: Naive Band Selection and Spectral Response Function (SRF) grouping, to align high-dimensional HSI inputs with the model's Sentinel-2 pretraining distribution. The experimental workflow of our research is presented in Figure \ref{fig:vis}. Our contributions are threefold. First, we present a baseline evaluation of the TerraMind GFM on four HSI-specific downstream tasks. Second, we perform a comparative analysis of channel sampling approaches for adapting HSI data to non-HSI architectures. Finally, we assess the potential of multimodal GFMs for spectral tasks by benchmarking our results against the HSI-native SpectralEarth model. 
\vspace{-5pt}
\section{Experimental Setup} \label{ExpSet}
\paragraph{Downstream Datasets.} We evaluate performance on four HSI benchmarks selected to represent a range of spectral difficulty. We define `spectral complexity' both by class granularity, ranging from broad land-cover mapping to fine-grained tree species identification, and empirically, based on the performance gap observed when reducing full hyperspectral inputs to 12 multispectral bands. Three datasets are from SpectralEarth \citep{Braham_2024}, aligning 202-band EnMAP imagery ($128 \times 128$ px chips) with ground truth masks, and the fourth is derived from the AI4EO challenge \citep{Nalepa_2024}. \textbf{EnMAP-BNETD} (2,100 chips, 10 classes) is a broad land cover task, where multispectral resolution is generally sufficient for discrimination. \textbf{EnMAP-CDL} (1,600 chips, 14 classes) presents a moderately complex crop segmentation task. \textbf{EnMAP-BDForet} (2,550 chips, 12 species) is a fine-grained tree species segmentation task. Distinguishing spectrally similar subtypes requires high spectral resolution, posing a strict challenge for non-HSI GFMs. \textbf{Hyperview-1} (1,732 train/1,154 test chips, 150 bands) is a scene-level regression dataset of four soil parameters: Potassium $K$, Phosphorus pentoxide $P_2O_5$, Magnesium $Mg$, and acidity $pH$, representing the highest spectral complexity. \looseness=-1

We report the mean Intersection over Union (mIoU) for the three segmentation tasks. For Hyperview-1, we use the challenge's normalized MSE: $\text{MSE}_{\text{norm}} = \sum_{i}^4 (\text{MSE}_{i}/\text{MSE}_{i}^{\text{base}})$, where $\text{MSE}_{i}^{\text{base}}$ is the error of a trivial mean-predictor on the training set.

\paragraph{TerraMind \& Channel Adaptation.}
TerraMind is a multimodal GFM pretrained on multispectral, SAR, Digital Terrain Model (DEM), and Land Use/Land Cover (LULC) data, along with derived spectral index layers. The optical component of the pretraining dataset consists of Sentinel-2 data (both L2A and L1C). To leverage the pretrained weights, we project the input HSI ($X_{\text{HSI}} \in \mathbb{R}^{H \times W \times C_{\text{in}}}$) into the S2L2A spectral space. We investigate two projection methods to obtain the adapted input $\hat{X} \in \mathbb{R}^{H \times W \times 12}$.

\textbf{Naive Band Selection:} We select the HSI bands centered closest to the nominal center wavelengths of the Sentinel-2 bands. Let $\Lambda_{\text{HSI}} = \{\lambda_1, \dots, \lambda_{C_{\text{in}}}\}$ be the center wavelengths of the HSI sensor, and $\Lambda_{\text{S2}} = \{\mu_1, \dots, \mu_{12}\}$ be the target S2 wavelengths. For each S2 band $k$, we select:
    \begin{equation}
        \hat{X}_{:,:,k} = X_{\text{HSI}} \left[:,:, \operatorname*{argmin}_{j} |\lambda_j - \mu_k| \right]
    \end{equation}

as the feature map. This method preserves the raw radiometric values of specific narrow bands but discards information from the rest of the spectrum.

\textbf{SRF-based Spectral Resampling:} To simulate a physically realistic S2 signal, we apply the Sentinel-2 SRF, mathematically following the spectral simulation approach in \citet{Aviles_2025a}. This operation is implemented as a fixed linear transformation (weighted sum) along the spectral dimension. Let $\phi_k(\lambda)$ be the continuous SRF for Sentinel-2 band $k$, and $\lambda_j$ be the center wavelength of HSI band $j$. We construct a weight matrix $\mathbf{W} \in \mathbb{R}^{C_{in} \times 12}$ where the unnormalized weights correspond to the SRF sensitivity at each HSI band center. To preserve radiometric consistency, we normalize the weights for each S2 band $k$ to sum to unity: $w_{j, k} = \phi_k(\lambda_j)$ and $\hat{w}_{j, k} = w_{j, k} / \sum_{m=1}^{C_{in}} w_{m, k}$. 

The final matrix $\widehat{\mathbf{W}} = \{\hat{w}_{j,k}\}$ is then applied to the input HSI data $X_{\text{HSI}} \in \mathbb{R}^{H \times W \times C_{in}}$ via a matrix-vector product along the spectral dimension, which aggregates information from all HSI bands falling within the S2 spectral range, providing a smoother, physics-aware representation:
    \begin{equation}
    \hat{X}_{h, w, :} = X_{\text{HSI}_{h,w,:}} \widehat{\mathbf{W}}, \quad \forall h, w.
    \end{equation}

\paragraph{Implementation Details.}
We fine-tune the \texttt{terramind\_v1\_base} backbone, initializing the optical modality with S2 weights. For comparability, we closely follow SpectralEarth's implementation setup. All experiments were implemented using the \texttt{terratorch} library \citep{Gomes_2025a}, which streamlines the adaptation of TerraMind to specific downstream tasks through its modular framework. We employ a Fully Convolutional decoder (256 channels) optimized using cross-entropy loss in the segmentation tasks. For regression, we attach a linear head (hidden dimension 256) optimized via MSE loss. All models are trained for 100 epochs using the AdamW optimizer, with early stopping (patience: 20). We utilize a Cosine Annealing scheduler for segmentation and ReduceOnPlateau for regression. To ensure statistical robustness, all experiments were repeated 10 times with different random seeds. Experiments were conducted on NVIDIA A100 SXM4 GPUs.

\section{Empirical Findings} \label{EmpFin}
Table \ref{tab:results} summarizes the results of our channel adaptation experiments, along with the HSI-native baselines reported in \citet{Braham_2024}. Qualitative results of the experiments with a short analysis can be found in the Appendix \ref{Appendix}. 
We observe two consistent patterns across the four datasets: \textbf{First}, the Naive Band Selection strategy consistently yields superior performance compared to the physics-aware SRF Grouping. In segmentation tasks, Naive Selection offers a gain of approximately +0.4\% to +3.4\% mIoU, while in the Hyperview-1 regression task, it achieves a significantly higher leaderboard rank (\#6) compared to SRF (\#25). 
This significant performance difference between the two band selection methods can suggest that TerraMind's pretraining has induced a strong sensitivity to specific spectral anchors corresponding to the Sentinel-2 center wavelengths. Naive selection preserves the raw radiometric distribution of these central wavelengths, making it a better simulation of S2 data for TerraMind. In contrast, SRF grouping creates a weighted-average signal along the spectral dimension. This acts as a 1D spectral low-pass filter, smoothing out sharp, narrow-band absorption features that are critical for discrimination. While physically more representative of the sensor properties, they represented a shift in TerraMinds learned representations.  
\textbf{Second}, comparing TerraMind to the HSI-native SpectralEarth baseline reveals a clear correlation between the performance gap and the spectral complexity of the task. On EnMAP-BNETD, the gap is small ($\sim$3\% mIoU), indicating that the task is ``spectrally easy'' and the model's pretrained spatial features are sufficient to compensate for the reduced spectral resolution (12 vs. 202 bands). However, on the ``Moderate'' CDL and ``Hard'' BDForet datasets, which require distinguishing between spectrally similar agricultural classes or tree species, the performance gap widens to $\sim$8\% and $\sim$11\%, respectively. This confirms that for fine-grained classification, the 12-band approximation is insufficient to capture the nuanced spectral signatures present in the full HSI input, regardless of the GFM's spatial priors. Notably, the lower absolute mIoU on the `easy' task results from its dense annotations, which penalize spatial errors more strictly than the sparsely annotated harder tasks (see Appendix \ref{Appendix}).
\begin{table}[h]
\centering
\caption{Comparative performance on downstream tasks (mean $\pm$ std over 10 runs). Metrics are mIoU for segmentation (higher is better) and Normalized MSE for regression (lower is better). \textbf{Bold} indicates the best performing model; \underline{underline} indicates the best adaptation strategy for TerraMind.}
\label{tab:results}
\resizebox{\columnwidth}{!}{%
    \begin{tabular}{ll c c c c}
    \toprule
     & & \multicolumn{3}{c}{\textbf{Segmentation Tasks (mIoU $\uparrow$)}} & \textbf{Regression (nMSE $\downarrow$)} \\
    \cmidrule(lr){3-5} \cmidrule(lr){6-6}
    \textbf{Model} & \textbf{Adaptation} & \textbf{EnMAP-BNETD} & \textbf{EnMAP-CDL} & \textbf{EnMAP-BDForet} & \textbf{Hyperview-1} \\
     & & \textit{(Easy)} & \textit{(Moderate)} & \textit{(Hard)} & \textit{(V. Hard)} \\
    \midrule 
    \multirow{2}{*}{\textbf{TerraMind}} & Naive Selection & \underline{0.465 $\pm$ 0.002} & \underline{0.693 $\pm$ 0.006} & \underline{0.657 $\pm$ 0.007} & \underline{0.813} (Rank \#6) \\[1ex]
     & SRF Grouping & 0.461 $\pm$ 0.003 & 0.679 $\pm$ 0.006 & 0.623 $\pm$ 0.006 & 0.831 (Rank \#25) \\
    \midrule\midrule
    \textbf{SpectralEarth} & \multirow{2}*{None (Full HSI)} & \multirow{2}*{\textbf{0.495 $\pm$ 0.001}} & \multirow{2}*{\textbf{0.774 $\pm$ 0.003}} & \multirow{2}*{\textbf{0.766 $\pm$ 0.005}} & \multirow{2}*{\textbf{0.810} (Rank \#5)} \\
    \textit{(Upper Bound)} & & & & & \\
    \bottomrule
    \end{tabular}%
}
\end{table}

Surprisingly, on the Hyperview-1 soil parameter estimation task, defined as the most spectrally challenging dataset, the Naive Selection approach with TerraMind performs competitively with the specialized SpectralEarth (0.813 vs 0.810 normalized MSE). This result indicates that the spatial representations learned by TerraMind can compensate for reduced spectral resolution, but it could also point to significant spectral redundancy in the task itself. Research in soil spectroscopy demonstrates that nutrients like $P, K$, and $Mg$ can be detected indirectly via correlations with primary constituents like Organic Matter (OM) and clay minerals \citep{Bogrekci_2005, Dematte_2017}. These constituents exhibit broad spectral responses that align closely with Sentinel-2's bands. Specifically, OM in the visible/red-edge and clay minerals in the SWIR (centered near 2200 nm). Consequently, the Naive Selection effectively captures these proxy signals while filtering out the noise inherent in the full hyperspectral continuum, allowing the TerraMind to perform effectively despite discarding $>90\%$ of the bands. Finally, the very small performance gap between TerraMind and SpectralEarth suggests a limit to performance on this dataset inherent to the task difficulty or label noise. This limit cannot be easily overcome with increased spectral resolution for generalist foundation models, implying that such problems demand more specialized modeling approaches. 

\section{Discussion \& Conclusion} \label{DiscCons}
This study investigated adapting multimodal GFMs not pretrained on hyperspectral data to hyperspectral downstream tasks. We did so by examining two techniques of HSI channel sampling. The results of our preliminary research on the integration of HSI into a multimodal GFM aim to answer the research question: Can a non-HSI GFM serve as an effective baseline for HSI-specific tasks? \looseness=-1 

Our findings suggest TerraMind produces a competitive baseline when the downstream task prioritizes spatial semantics over spectral precision. On EnMAP-BNETD, the model reached within 3\% of the HSI-native SpectralEarth model, leveraging pre-learned spatial representations to compensate for the massive spectral reduction (202 $\to$ 12 bands). However, for tasks requiring fine-grained spectral discrimination, such as EnMAP-BDForet, the ``spectral gap'' cannot be overcome via simple subsampling. This highlights that while existing multimodal GFMs are powerful generalist tools, they cannot replace specialized HSI architectures for tasks where precise spectral information guides prediction. 
Counter-intuitively, Naive Band Selection consistently outperformed physics-based SRF Grouping. We attribute this to how TerraMind anchors its representations to specific S2L2A bands, suggesting this may be a model-specific result that necessitates validation through similar experiments on other multimodal GFMs. Alternatively, the performance drop suggests that SRF sampling acts as a spectral low-pass filter \citep{Aviles_2025a}, smoothing out precise spectral features and thereby reducing their informativeness for class discrimination.

These preliminary results underscore the need to move beyond simple adaptation toward full integration. Future research will extend benchmarks to other spectrally intensive applications and sensor modalities (e.g., PRISMA, EMIT). Methodologically, we plan to integrate native HSI support within TerraMind via a hyperspectral tokenizer to ingest full-spectrum data. Finally, to fully isolate the contribution of learned spatial priors, future work will incorporate stronger baselines and empirically validate the model's representation geometry.

\section*{Acknowledgements}
This study was funded by the Italian Space Agency, ASI, under convention n. 2024-28-HH.0 - CUP n. F43D24000110001.

We acknowledge ISCRA for awarding this project access to the LEONARDO supercomputer, owned by the EuroHPC Joint Undertaking, hosted by CINECA (Italy).

Gemini 3 was used to improve the manuscript's language.

\bibliography{iclr2026_conference}
\bibliographystyle{iclr2026_conference}

\newpage
\appendix
\section{Appendix} \label{Appendix}
\begin{figure}[h]
    \centering
    \includegraphics[width=\linewidth]{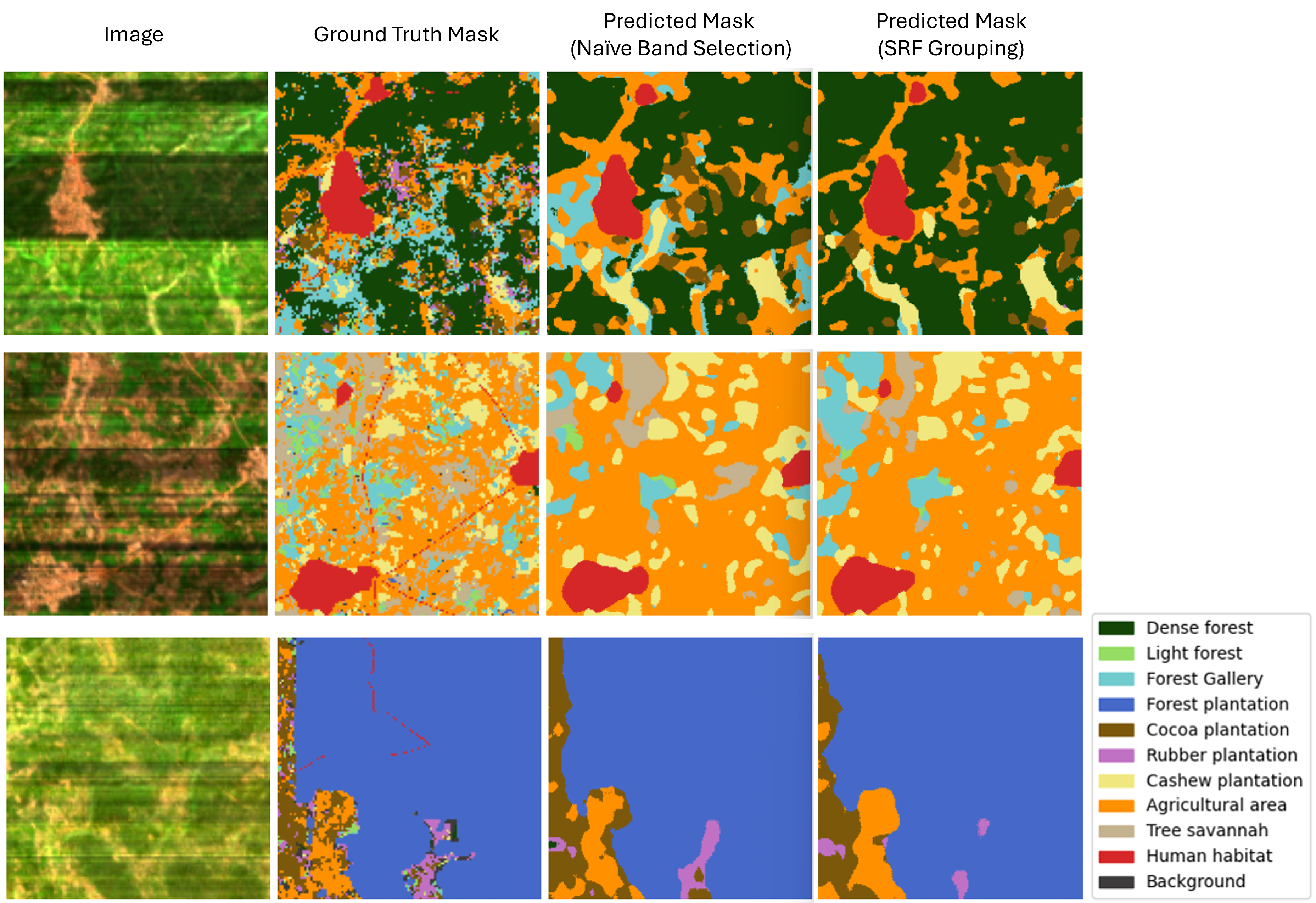}
    \caption{Prediction examples of TerraMind on the \textit{EnMAP-BNETD} dataset. Compared to SRF Grouping (right), Naive Band Selection (center) better preserves fine spatial structures.}
    \label{fig:BNETD_preds}
\end{figure}

\begin{figure}[h]
    \centering
    \includegraphics[width=\linewidth]{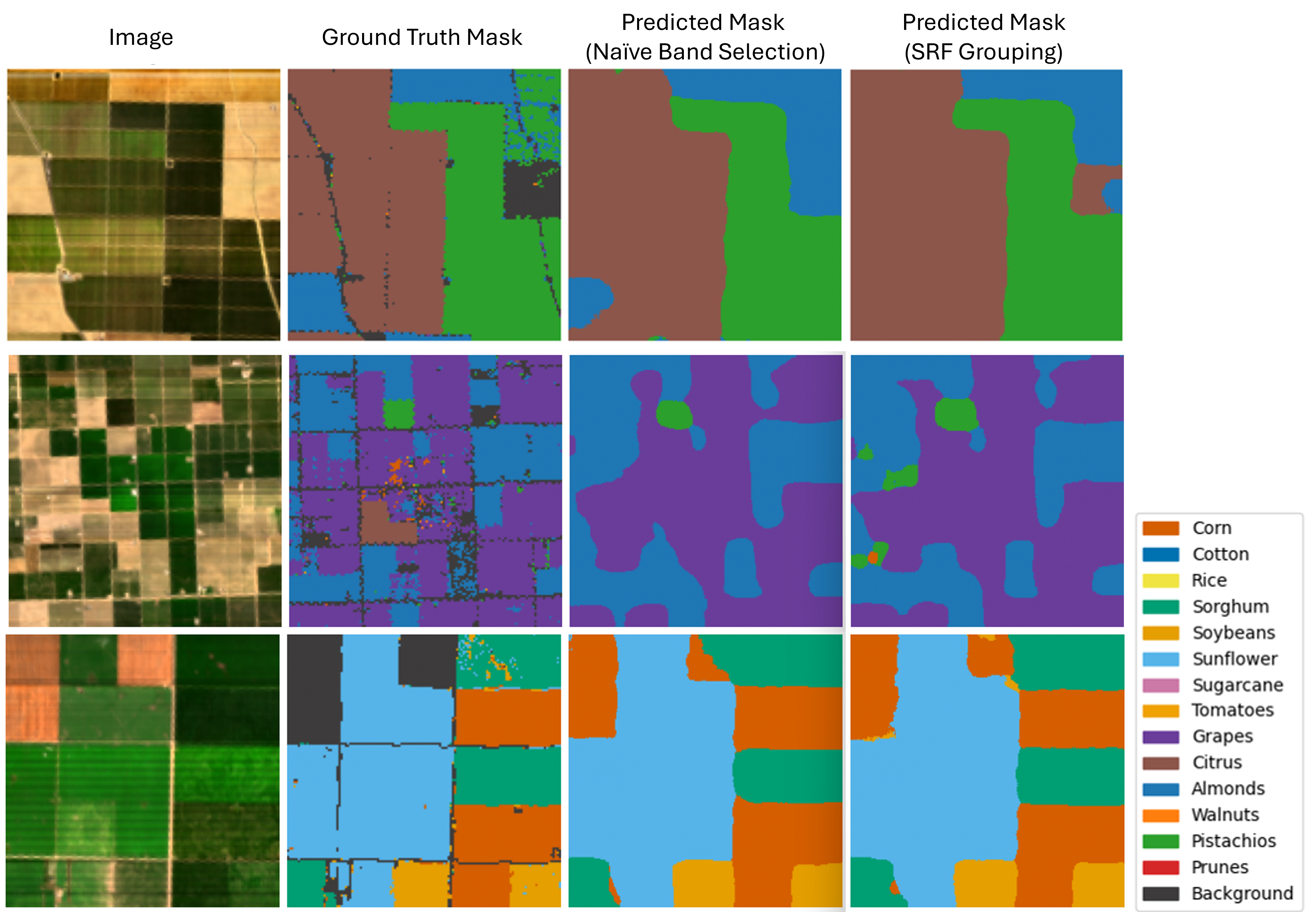}
    \caption{Prediction examples of TerraMind on the \textit{EnMAP-CDL} dataset using both band sampling techniques. Note that the background class was set as the ignore index during training.}
    \label{fig:CDL_preds}
\end{figure}

\begin{figure}[h]
    \centering
    \includegraphics[width=\linewidth]{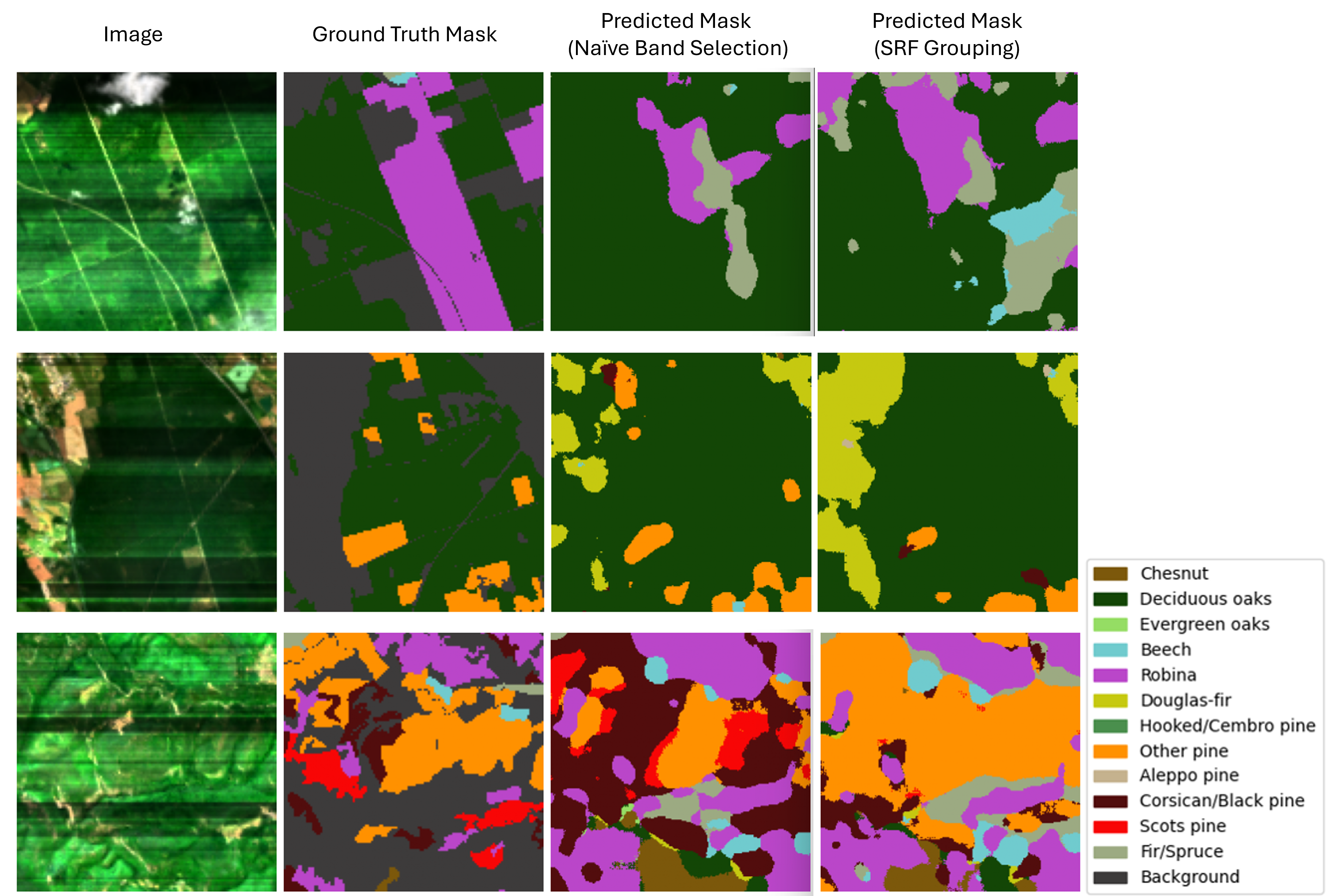}
    \caption{Prediction examples on \textit{EnMAP-BDFORET}. The spectral similarity of tree species leads to visible class confusion in the SRF predictions (right), whereas Naive Band Selection (center) achieves better discrimination between pine subtypes. Note the extensive background regions (grey), which were set as the ignore index during training.}
    \label{fig:BDFORET_preds}
\end{figure}

Figures \ref{fig:BNETD_preds} through \ref{fig:BDFORET_preds} display sample predictions from the fine-tuned TerraMind model on the segmentation datasets.
The predictions confirm the findings in Section \ref{EmpFin}. First, performance degrades on spectrally complex tasks, with significant class confusion evident in Figure \ref{fig:BDFORET_preds}, where distinct tree species are mixed up due to insufficient spectral detail. The qualitative difference between adaptation strategies is also evident. In the first example of Figure \ref{fig:BNETD_preds}, the SRF Grouping method fails to detect the `Forest Gallery' class and produces over-smoothed features that do not match the fragmented Ground Truth. In contrast, Naive Band Selection better preserves these fine spatial structures (Figure \ref{fig:BNETD_preds}, last row) and maintains better separation between pine classes (Figure \ref{fig:BDFORET_preds}). 

We attribute the general coarseness of the spatial boundaries to the use of a standard $16\times16$ patch size, as supported by findings in \citet{Braham_2024}. 

Finally, we note that the sparse annotations in CDL and BDForet result in the model predicting valid classes within `ignore index' (background) regions. This sparsity masks spatial errors on the harder tasks, inflating their metrics compared to the more densely annotated BNETD, where every boundary mistake is penalized.

\end{document}